%% file: main.tex
\documentclass[letterpaper, 10 pt, conference]{ieeeconf}  % Comment this line out if you need a4paper

\IEEEoverridecommandlockouts                              % This command is only needed if 
\usepackage{graphics} % for pdf, bitmapped graphics files
\usepackage{graphicx}
\usepackage{color} 
\usepackage{subfigure}
\definecolor{listgray}{rgb}{0.88,0.88,0.88} 
\usepackage{listings}
\lstdefinestyle{yaml_listing}{
  language=Python,
  numbers=none,
  stepnumber=1,
  tabsize=2,
  showspaces=false,
  showstringspaces=false,
  float = [b],
  backgroundcolor=\color{listgray}, 
  captionpos=b, 
  basicstyle=\footnotesize, 
  frame=tbrl, %t: top, r, b, l 
}
\newsavebox{\mybox}
\usepackage{amsmath} % assumes amsmath package installed
\usepackage{amssymb}  % assumes amsmath package installed
\usepackage{hyperref}

\title{\LARGE \bf
ARChemist: Autonomous Robotic Chemistry System Architecture*
}

\author{Hatem Fakhruldeen$^{1}$, Gabriella Pizzuto$^{1}$, Jakub Glowacki$^{1}$ and Andrew Ian Cooper$^{1}$% <-this % stops a space
\thanks{*This work was supported by the Leverhulme Trust through the Leverhulme Research Centre for Functional Materials Design and the H2020 ERC Synergy Grant \textit{Autonomous Discovery of Advanced Materials} under grant agreement no. 856405.}% <-this % stops a space
\thanks{$^{1}$Hatem Fakhruldeen, Gabriella Pizzuto, Jakub Glowacki and Andrew Ian Cooper are with the Leverhulme Research Centre for Functional Materials Design,
        University of Liverpool, United Kingdom
        {\tt\small h.fakhruldeen@liverpool.ac.uk}}%%
}

\begin{document}

\maketitle
\thispagestyle{empty}
\pagestyle{empty}

%%%%%%%%%%%%%%%%%%%%%%%%%%%%%%%%%%%%%%%%%%%%%%%%%%%%%%%%%%%%%%%%%%%%%%%%%%%%%%%%
\begin{abstract}

Automated laboratory experiments have the potential to propel new discoveries, while increasing reproducibility and improving scientists' safety when handling dangerous materials.  
However, many automated laboratory workflows have not fully leveraged the remarkable advancements in robotics and digital lab equipment.
As a result, most robotic systems used in the labs are programmed specifically for a single experiment, often relying on proprietary architectures or using unconventional hardware.
In this work, we tackle this problem by proposing a novel robotic system architecture specifically designed with and for chemists, which allows the scientist to easily reconfigure their setup for new experiments.
Specifically, the system's strength is its ability to combine together heterogeneous robotic platforms with standard laboratory equipment to create different experimental setups.
Finally, we show how the architecture can be used for specific laboratory experiments through case studies such as solubility screening and crystallisation. 

\end{abstract}

% Fabricated cellular micro-scaffold that recapitulates natural extracellular matrix (ECM) has shown huge potential in the study of cell behaviors. However, the reproducing of the physiological morphology with high efficiency and accuracy in the micro-scaffold still remains as a major challenge. Here, we propose a novel automated fabrication method to engineer high-fidelity cellular micro-scaffold with a proportion-corrective control algorithm to modulate the photocuring process of biodegradable hydrogel in real-time. A digital holographic microscopy (DHM) system is integrated into the micro-fabrication system based on the digital micro-mirror device (DMD) to enable the real-time detection of the photocuring process. Before the photocuring, the theoretical curing thickness is determined by the calibrated model. To fabricate a micro-scaffold with high-fidelity morphology, the incident UV light is divided into different grid areas and achieve local discrete photocuring control. For every local area, the real-time added value of the cured thickness is compared with the theoretical value to determine the distortion which is corrected by the second-step exposure controlled by the proportion-corrective algorithm. Finally, the algorithm efficiently improved the fabrication accuracy from 200μm to 50μm. With the long-term culture, the cells viabilities exceeded 96%. The experimental results verified the effectiveness and feasibility of the proposed control algorithm.

%%%%%%%%%%%%%%%%%%%%%%%%%%%%%%%%%%%%%%%%%%%%%%%%%%%%%%%%%%%%%%%%%%%%%%%%%%%%%%%%
\input{section/introduction}
\input{section/related_work}
\input{section/architecture}
\input{section/experimental_evaluation}
\input{section/conclusion}
\section*{ACKNOWLEDGEMENTS}
The authors thank Louis Longley for his assistance with the case studies and the Cooper group for feedback on the chemical recipe.

% \begin{thebibliography}{99}

% \bibitem{c1} X

% \end{thebibliography}

\bibliographystyle{IEEEtran}
\bibliography{references}

\end{document}

%% file: section/introduction.tex
\section{INTRODUCTION}
\label{sec:introduction}

Accelerating the discovery of new materials is important for industrial applications such as healthcare and energy production. 
This can be achieved through running long-term experiments autonomously, for example by increasing the use of robotic platforms in laboratories. 
In practice, this would accumulate more experiments in less time, and potentially minimise the scientists' exposure to harmful chemicals, reducing their repetitive tasks.

Recently, there have been significant efforts to fully automate laboratory experiments with the introduction of different robotic platforms such as for the search of materials for hydrogen production \cite{Burger2020} and for solubility screening \cite{Hein2021}. 
For example, in \cite{Burger2020} the robot ran for eight consecutive days, carrying out 688 experiments, which is notably faster than a human scientist. 

However, while these have been important achievements, there is still an open gap when it comes to having a standard architecture that allows ease of integration of robotic platforms and laboratory equipment simultaneously, but that is also easy to use by chemists.
Being (in general) non-proficient robotic users, chemists tend to prefer a system that would be easy to program, simplifies experimental data analysis, and generalises to a broad array of experiments. 

\begin{figure}[!t]
  \includegraphics[width=0.45\textwidth]{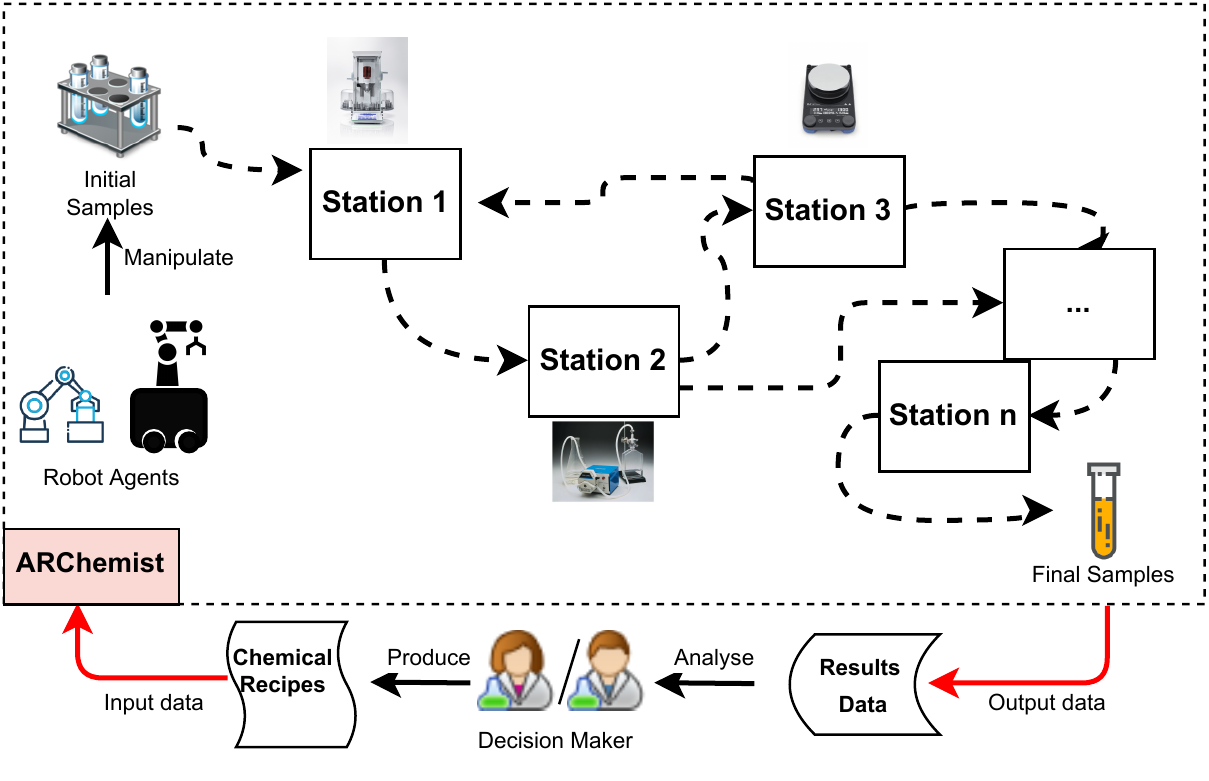}
  \caption{Scheme for ARChemist. The overall reconfigurable system allows chemists to program their experiments through a chemical recipe, capturing the workflow state in terms of stations, robots, and sample progress. The chemical recipes represent the system input data, while the fully processed samples results represent its output data.
}
  \label{image:experimental_setup}
\end{figure}

To address this, we propose a novel architecture specialised in running standard chemistry laboratory routines using heterogeneous collaborative robotic platforms and various lab equipment (Fig.~\ref{image:experimental_setup}). 
The main advantage of our architecture is that it streamlines the process of automating chemistry experiments by granting the scientist the possibility to easily design and execute his or her experiment. 
We achieved this through designing the architecture in close collaboration with chemists: as a result, we provide the user an input file to program the overall experiment.
Our heterogeneous robotic architecture is evaluated on two essential use cases: solubility screening and crystallisation. 

In summary, the contributions of this work are as follows: 

\begin{enumerate}
\item A novel reconfigurable architecture using heterogeneous robots and laboratory instruments that can be easily expanded with additional equipment through an open-source middleware. Specifically, this architecture shows the potential of using different robotic platforms in the chemistry automation domain, where the scientist can build his or her experiment using a human-readable language.

\item Real-world examples of laboratory experiments such as solubility screening as well as crystallisation and analysis using different robots and laboratory equipment. This is realised by creating synergy between lab hardware and `robotic chemists' through a communication layer.

\end{enumerate}

% This rest of the paper is structured as follows. 
% In Section~\ref{sec:related_work}, we give an overview of related work in laboratory automation, whilst outlining how ARChemist addresses the open gaps.
% Section~\ref{sec:archemist_description} reports the overall system architecture and Section~\ref{sec:experimental_evaluation} details the experimental evaluation of typical case studies in chemistry lab automation. 
% Finally, Section~\ref{sec:conclusion} draws conclusions and gives direction for future work. 

%% file: section/related_work.tex
\section{RELATED WORK}
\label{sec:related_work}

Robots are increasingly being used in laboratories to ensure the safety of scientists by mitigating their exposure to toxic and strenuous environments, while preserving their overall health with fewer repetitive tasks.
The overarching goal has always pivoted around the importance of developing architectures that increase reproducibility and throughput.

% In the life sciences, one of the earlier works~\cite{Fleischer2018} focused on automating the process of acquiring analytical measurements in sample preparation. 
% In comparison to a scientist, robots offer increased repeatability and precision of such repetitive tasks. 
% Given that the use and development of `robotic scientists' is still in its infancy, the research efforts towards unifying robots and lab instruments are relatively small. 
% In~\cite{Neubert2017}, the authors propose 
% a hierarchical workflow management system that integrates mobile
% robots and mobile devices through the instrument layer, to bring together the scientist and the equipment in an automated or semi-automated manner. 
% Nonetheless, the architecture does not include heterogeneous robotic platforms and seems to be developed mainly for the life sciences. 

Chemistry is a scientific field that has gained interest in using robotic platforms for automating experiments.
A mobile robotic chemist was used to search for materials, namely photocatalysts, for hydrogen production from water~\cite{Burger2020}. 
The robot carried out experiments to validate several hypotheses, accelerating the work of a human researcher, which would have taken several months rather than only a week. 
This work aligns with our architecture, such that the goal of our method is to automate the scientists rather than the instruments.
Our work plans on augmenting the capabilities of a mobile robotic platform, as used in this earlier work, with multiple platforms that would also generalise to more experiments beyond the search of materials for hydrogen production.
Furthermore, these previous experiments were carried out using a rigid configuration and hence would not allow migration with ease to a new workflow.
In contrast, our new work proposes an architecture that uses the same mobile manipulator, together with other robots and lab equipment, yet can be easily reconfigured to develop new workflows and add additional platforms. 

In the field of automated synthesis, previous studies proposed a preliminary integration of algorithmic prediction of viable routes to a desired compound with its implementation on a platform that needs little to no human intervention~\cite{Coley2019}. 
However, human intervention was still necessary to aide the predictor with practical considerations in for example solvent choice.
A further work in the field of organic chemistry synthesis used a robot that was able to perform a Michael reaction, and the results obtained were comparable to a junior chemist~\cite{Lim2020}. 
The setup was optimised for a single reaction and there was no feedback between the instruments and the robot; two issues that our architecture aims to address.

Another prominent work is the chemical processing programming architecture (ChemPU)~\cite{Hammer2021}, which is a more generalisable programming architecture with customised hardware that aims to standardise all chemical synthesis. 
Here, the authors propose an architecture for several chemistry automation experiments aimed at increasing reproducibility and interoperability. 
However, the proposed architecture requires the ChemPU and cannot easily be integrated with other standard equipment found in chemistry labs.
Also, neither of these approaches integrates mobile robots.
The latter is a key requirement as most scientific labs have been designed for human scientists and as a result, it is often the case that other equipment are not within reach of the robot.
Hence, mobility is essential to transport samples to other stations.

Another recent work developed a robotic system for automated solubility screening~\cite{Hein2021}, a vital step in many chemistry experiments.
The authors combine solid and liquid dispensing with computer vision and iterative feedback to measure caffeine solubility. 
Although this system requires initial user input, it is then capable of running autonomously and producing results which are comparable to other invasive techniques like for example high-performance liquid chromatography.
In their work, the authors demonstrate their methodology with both the N9 and the Kinova Gen3 robotic arms separately. 
While our work also uses solubility screening as a case study, we extend the workflow by adding a mobile manipulator, which would both facilitate transporting samples between isolated lab regions, and could potentially also work in parallel to the manipulator on other stations with the goal of increasing throughput. 

Our proposed architecture differs from earlier related works~\cite{Hein2021}~\cite{Fleischer2018} by using a widely accepted open source middleware that would ease integration with additional robotic platforms and laboratory equipment. 
The main motivation for developing our architecture was to move closer to having reliable and robust `robotic chemists'; this needs to be an interdisciplinary effort and we believe that using the robotic operating system as a middleware would fuel further interest from the robotics community.
Moreover, to our knowledge, our work is the first to introduce an architecture for lab automation that uses heterogeneous robotic platforms within the same experimental workflow.
As a result, our work aims to illustrate the use of multiple robotic platforms within chemistry workflows, to increase both output and allow generalisation to various experiments.

%% file: section/architecture.tex
\section{ARCHEMIST DESCRIPTION}
\label{sec:archemist_description}
\begin{figure*}[!t]
  \centering
  \includegraphics[width=0.7\textwidth]{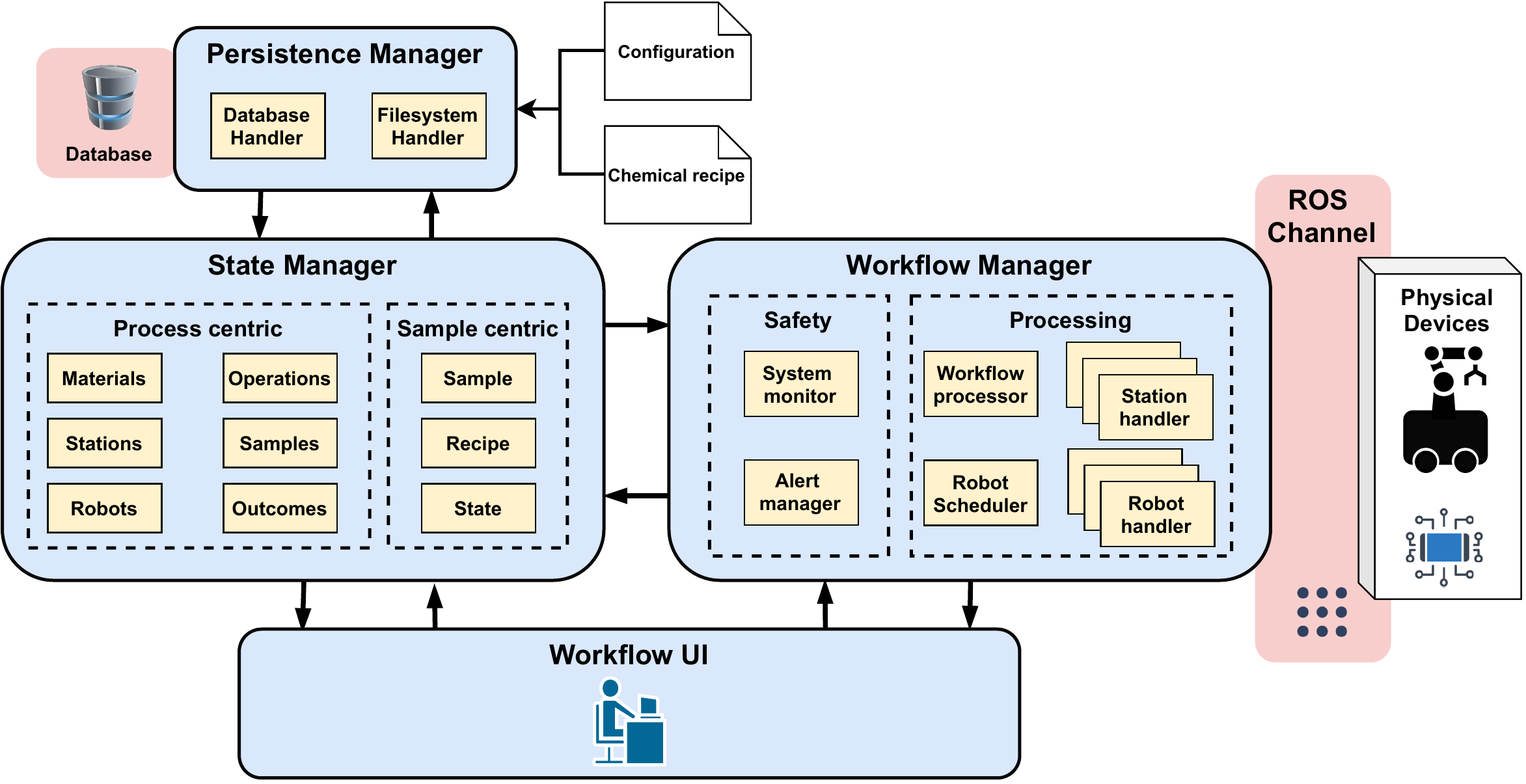}
  \caption{An overall block diagram of ARChemist. The State Manager is responsible for representing the chemistry experiment state. The Persistence Manager allows to store and retrieve this state from a database. In addition, it is responsible for parsing the input files. The Workflow Manager utilises the previous modules to process the experiment samples and assign them to their respective robots and stations. The architecture utilises ROS as a communication layer to interact with the physical robots and lab equipment. The Workflow User Interface allows the scientist to interact with the system and provide their input.}
  \label{img:arch_blockdiagram}
\end{figure*}

\subsection{Overview}
\label{subsec:description_overview}
% can it be effortless design of chemistry workflow and autonomous execution???
The ARChemist system architecture (Fig.~\ref{img:arch_blockdiagram}) allows us to combine heterogeneous robotic platforms and lab instruments efficiently for autonomous chemistry workflows. 
Using human-readable input chemistry recipes, scientists can easily program their experiments without the need for extensive robotic expertise.  
These recipes can be used by artificial or human agents to interact with the system and to make decisions to drive the chemical process towards a desirable outcome. 
This architecture is re-configurable and easily extended with new lab equipment and robots without changing its underlying structure or processing algorithm.\par

% For the ARChemist architecture to perform the chemistry experiment and execute the workflow, it uses the information stored in a chemical recipe, specially designed to represent the experiment workflow.
%in terms of materials, stations and workflow. 
% In this formulation, 
Each individual sample has a recipe attached to it, which needs to be fulfilled to complete the experiment. 
To achieve that, the experiment space is represented in terms of the available resources in the chemistry workflow, that include materials, robots, equipment, and samples. 
Moreover, it also keeps track of sample information e.g. materials content and operation history. 
This is managed and tracked by the State Manager, which initialises the system state from a configuration file that describes the available resources or from database records, if this is not the first run. 
The database and file handling are managed by the Persistence Manager. 
It allows storage of the system state in a database for easy retrieval. Furthermore, it also handles parsing configuration and recipe files and constructing their respective objects.\par

\begin{figure}[!t]
  \centering
  \includegraphics[width=0.45\textwidth]{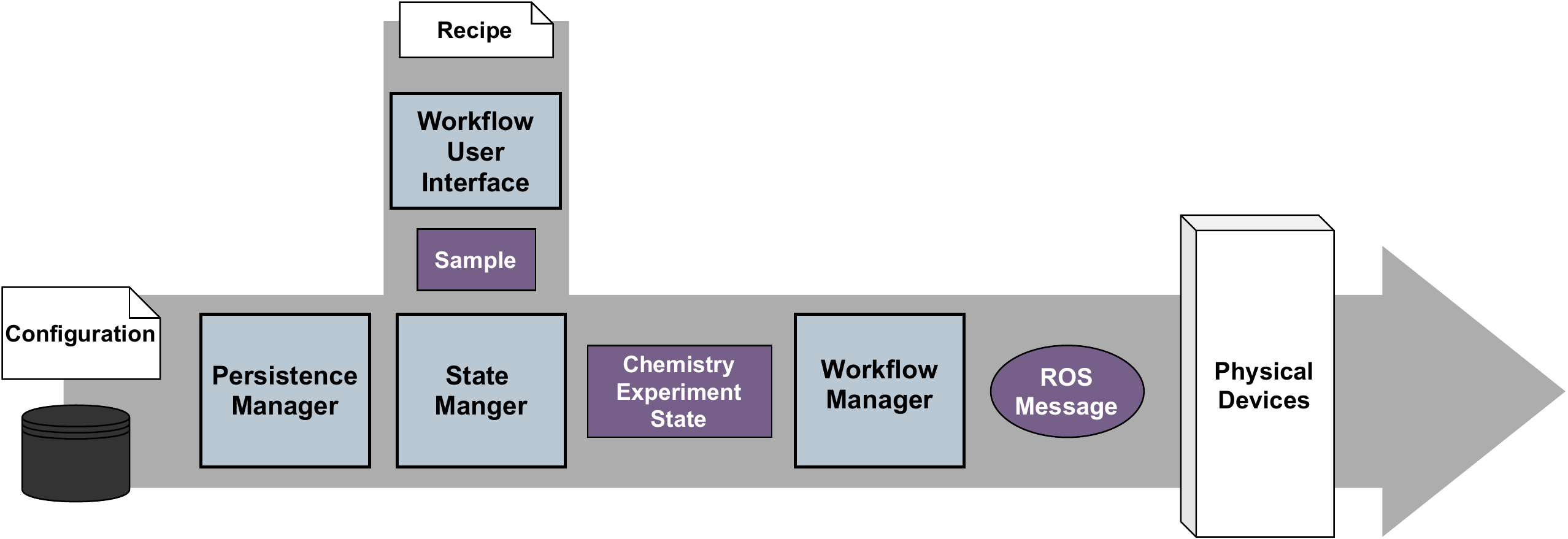}
  \caption{An abstract representation of the ARChemist architecture data flow between its various elements.}
  \label{img:arch_dataflow}
\end{figure}

The experiment execution and monitoring are handled by the Workflow Manager, which utilises the state representation provided by the State and Persistence Managers to perform its duties. 
To process the samples and execute their chemical recipes, the Workflow Processor deals with any samples not assigned to stations and according to the state and recipe either assigns them to a station or adds them to a robot job queue. 
The Robot Scheduler handles the job queue and assigns a sample for a robot to manipulate according to its location and capabilities. Station and Robot Handlers run as parallel processes and start processing samples when they get assigned to their respective stations. These handlers use the appropriate ROS~\cite{Quigley09} channels to communicate with their respective robots/stations. On the other hand, the System Monitor observes the operational and safety status of the different robots/stations used in the workflow and would halt the system if critical failures happen. At the same time, the Alert Manager utilises a number of user-defined rules to monitor the state of the chemistry experiment and issues alerts when rules are violated. Finally, the Workflow User Interface allows the scientist to interact with the system by adding a fresh sample and its attached recipe so that the system can start processing it.
The overall dataflow in the architecture is illustrated in Fig~\ref{img:arch_dataflow}.\par

\subsection{State Manager}
\label{subsec:state_manager}

The State Manager represents, stores and updates the space of the chemistry experiment. 
It represents the system state in terms of the available material, robots, lab instruments, operations and samples. 
In combination with the Persistence Manager, allows it to track the system state and record all the relevant information. Namely, it allows the other architecture components to retrieve the system state and update it as they see fit when performing their operations.\par

The chemistry experiment space can be represented in terms of process-centred and sample-centred parameters. The former describes all the resources available for the process or workflow. These include: i) chemical materials and their physical quantities, ii) available lab instruments/equipment and robotic platforms and iii) the samples circulating through the system for processing. On the other hand, sample-centred parameters include the individual samples, their respective information and associated chemical recipes. These parameters are further described as follows:

\subsubsection{Process-Centred Parameters}
\label{subsubsec:process_centred}
Materials are represented through their physical state, mass, density and volume. 
The lab instruments and robots are represented in terms of their locations in the environment, their operational status, availability for processing new batches and provided operations. To elaborate, these operations represent the actions that can be performed by the stations to process the samples and advance their state. For lab instruments, these can vary between altering the materials' state and measuring their various properties. 
At the same time, robots' operations include transporting and manipulating samples between stations. Consequently, each station/robot has a number of specific operation descriptors that are defined in terms of input parameters and their outcomes. These outcomes can be as a simple as Boolean values describing the success or failure of the operation, to more sophisticated ones describing measurement operations. These operation and outcome descriptors are used in the chemical recipe to describe the needed operations for the chemistry experiment. Moreover, they are used to log the sample history when travelling through the process and store its measurements. These parameters are initialised from a configuration file that describes all the experimental resources (stations, robots and materials). Alternatively, these parameters can also be initialised from the database if records from previous runs exist.\par

\subsubsection{Sample-Centred parameters}
\label{subsubsec:batch_centred}
Each sample has a chemical recipe associated with it that describes the required materials, stations' operations and process workflow between these stations in order to conduct the perspective chemistry experiment. Consequently, each sample is represented in terms of its materials content, their quantities, the history of operations it has undergone, their outcomes and current process status. This includes its physical location in the workspace, current progress through the workflow and its assignment status, \textit{i.e.}, if it is assigned to a station or robot. Samples alongside their attached recipes are added to the system via the Workflow User Interface for processing. 

\subsection{Workflow Manager}
\label{subsec:workflow_manager}
The Workflow Manager is responsible for the chemistry experiment execution, monitoring and safe operation of the architecture. It uses the State and Persistence Managers to retrieve the system state and update it after processing. It comprises number of sub-modules that can be grouped under process execution and safety monitoring categories:

\subsubsection{Workflow Processor}
\label{subsubsec:workflow_processor}
This sub-module is responsible for processing the samples progressing in the experiment and assign them an appropriate operation depending on their recipe.
This processor continuously checks for any samples not assigned to stations or robots in the system. Consequently, depending on their current workflow progress and recipe information, it assigns them to stations. If they are not physically located at their upcoming stations, it adds them to the robot job queue that is handled by the Robot Scheduler so they can be transported or manipulated to progress. When a station or a robot successfully processes their assigned sample, the state and any changes they have undergone are forwarded to the State Manager and consequently added to their processed sample list that the processor continuously monitors.

\subsubsection{Robot Scheduler}
\label{subsubsec:robot_scheduler}
The robot scheduler goes through the robot job queue and depending on the required robotic operation associated with the sample, it assigns it to the appropriate robot. In other words, it goes through all available robots and selects the most appropriate one for the task, depending on its capabilities and location.

\subsubsection{Station and Robot Handlers}
\label{subsubsec:handlers}
A handler is used for each station and robot utilised in the experiment. These handlers run as parallel processes that utilise the State and Persistence Managers to monitor their appropriate stations/robots' states for any newly assigned samples. Once a sample is assigned to a station or robot, the handler would retrieve the operation attached to it and start executing it using the ROS topics associated with its robot/instrument. Upon completion, it would update the sample and its own state before adding it to their processed samples list.

\subsubsection{System Monitor}
\label{subsubsec:system_monitor}
This sub-module continuously checks the availability and safety status of the available stations and robots and updates their states accordingly using the State Manager. The Workflow Processor utilises this information before assigning samples to stations and robots and would stop execution if any of them is not operational or has a safety stop.

\subsubsection{Alert Manager}
\label{subsubsec:system_monitor} 
It relies on user-defined rules to monitor various parameters in the system and depending on their severity, simply notify the user for a need of action or halt the whole process altogether. An Example of such rules including monitoring the chemistry process parameters such available quantities of liquids and solids available for the experiment and notify the user when they get below a certain threshold.

\subsection{Persistence Manager}
\label{subsec:persistence_manager}
The Persistence Manager stores the chemistry experiment state using a database and making it persists over multiple runs and to be accessible across the various architecture sub-modules. This manager allows different sub-modules running across different processes to access the world state and update their relevant information simultaneously. Moreover, it also handles reading and parsing configuration and recipe files to retrieve their content and construct the appropriate program objects. These  allow external decision makers to interact with the system and command its execution process.

\subsection{Workflow User Interface, ROS Channels and Physical Devices}
\label{subsec:physical_devices}

The Workflow User Interface allows the scientist to interact with the system, to add fresh batches and monitor the system execution state. Currently, this is done via a command line interface, which we plan on improving with a graphical user interface in the near future.\par

In our architecture, ROS is utilised as the communication layer between the architecture and the physical robots and lab instruments located in the experiment workspace. It was selected because of its reliability, ease of use and asynchronous nature. To integrate lab equipment with ROS, custom drivers needed to be developed to allow their integration with robots.
 
Whilst the architecture is flexible enough to easily expand to additional physical devices, here we used two widely available robotic platforms: Kuka KMR and Franka Emika Panda, and the following standard laboratory equipment: solid and liquid dispensers, balance, and hot plate and stirrer.

In practice, there are multiple instruments that would also be useful to integrate in a chemistry lab; hence, the robots and equipment presented here (Section~\ref{sec:experimental_evaluation}) are only detailed as these were used for our case studies.
Nonetheless, the overall architecture was developed such that other robotic platforms and equipment can be easily integrated and we plan on extending this as we get more feedback from chemists as they use this architecture for their work.

\subsection{Chemical Recipe}

To represent chemistry experiments, a special recipe representation for its operations and parameters was developed.
This representation was designed after going through the literature and with feedback obtained directly from chemists.
Here, chemistry experiments are human-readable, easily interpreted and concise enough to capture the details without over-generalising or being generic. 
It describes the experiment in terms of the resources and operations applied to the individual sample to process it in the experiment and achieve the desired outcome. This recipe description includes:

\begin{itemize}
    \item The quantities of materials added to the sample vial.
    \item The stations that the sample will go through in order to be processed and the operations involved there. These operations would be described in terms of the stations' specific actions, their input parameters and associated outcome descriptor that represent these actions results.
    \item The workflow between these station and their specific operations described as a state machine that allow non-sequential process execution.
\end{itemize}

\lstset{basicstyle=\ttfamily\srcsize,style=yaml_listing}
% \begin{resizebox}{0.5\textwidth}
%\begin{minipage}{.95\columnwidth}
\begin{center}
\begin{lrbox}{\mybox}
\begin{lstlisting}[language=python, caption={Simplified chemical recipe for dispensing solid (sodium chloride) and solvent (water) using the Quantos and peristaltic pumps.}, label={fig:recipe}]
chemical_recipe:
  name: sample_recipe
  materials:
    liquids: { water }
    solids: { NaCl }
  stations:
    solid_dispensing_quantos_QS2:
      stationOp:
        dispense_solid:
          properties:
            solid: NaCl
            mass: 15
            unit: mg
          output: 
            name: "final_weight" 
    peristaltic_liquid_dispensing:
      StationOp:
        dispense_liquid:
          properties:
            liquid: water
            volume: 2
            unit: mL
          output: 
            name: "dispensed_volume"
  stationFlow:
    start:
      onSuccess: solid_disp
      onFail: end
    solid_disp:
      station: "solid_dispensing_quantos_QS3"
      task: {"dispense_solid", NaCl, 15, "mg"}
      onsuccess: liquid_disp
      onfail: end
    liquid_disp:
      station: "peristaltic_liquid_dispensing"
      task: {"dispense_liquid", water, 2, "mL"}
      onSuccess: end
      onFail: end
    end:

\end{lstlisting}
\end{lrbox}
\scalebox{0.7}{\usebox{\mybox}}
\end{center}
% \end{resizebox}
This description builds on our state representation described in section~\ref{subsec:state_manager} of the chemistry experiment space. 
Listing~\ref{fig:recipe} illustrates an example recipe for a simplified material dispensing setup. 
YAML data serialisation language and its syntax were used to represent this recipe. 
This language was selected because of readability, ease of parsing and interpretation.  
The recipe describes the process of dispensing sodium chloride, followed by dispensing a solvent.

\subsection{Reconfiguration}

To facilitate the discovery nature of chemistry experiments, the ability to add new operations/stations to the workflow and remove them when needed in order to modify the process is required. Consequently, the ARChemist architecture was designed with this requirement such that there is a loose coupling between the station/operations and the processing Workflow Manager sub-modules. It allows to easily reconfigure the system according to the process requirements using a configuration file, which describes the stations, robots and operations involved in the process. The proposed architecture when initialising uses this file to load the required stations and start their respective handlers accordingly. This was achieved by using the dynamic typing and type-introspection features of the Python programming language, where after parsing the station name from the configuration file, it searches inside the defined modules for the classes representing these stations and/or robots.\par

By the same token, this allows the system to be easily extended and reusable in different contexts. To add new stations/robots, the user needs only add their respective class file and station handler to the appropriate module, without modifying any other parts of the system. Consequently, the system when initialising and loading the experiment stations/robots will discover these new stations and robots and start using them straight away. This is a powerful feature that allows the system to evolve and be highly customisable. \par

Finally, it is worth noting that the architecture allows the system to work with multiple different recipes, provided all their stations/operations were defined in the given configuration file.\par

%\end{minipage}

%% file: section/experimental_evaluation.tex
\section{EXPERIMENTAL EVALUATION}
\label{sec:experimental_evaluation}

For our experimental evaluation, after thorough discussions with chemists, we settled on two case studies: automated solubility screening (\ref{subsec:case_study_1}) and crystallisation (\ref{subsec:case_study_2}).

In both case studies, the KUKA KMR mobile manipulator and the Franka Emika Panda robot manipulator were used. 
Being a mobile manipulator, the KUKA KMR was mainly used to transport samples between stations found across different parts of the lab. 
% The robotic platform comprises of the KMP200 omniMove mobile base (with two SICK S300 safety laser scanners) and a LBR iiwa14 R820 seven degree-of-freedom robotic arm. 
The Franka Emika Panda, another seven degree-of-freedom robotic arm, was then mainly used for manipulating the vials in the different stations.

Most experiments in the field of material discovery start by dispensing a solid into a vial. 
The solid was accurately dispensed using the Quantos QS30 instrument (Mettler Toledo), which is widely used in the pharmaceutical industry. 
This is often followed with liquid dispensing.
In our case, a bespoke system built from a 200 series mini peristaltic pump was used~\cite{Burger2020}. 
This method dispenses liquids gravimetrically using a feedback loop used, which given the right form of tubing, can dispense a vast array of solvents. 
Following safety regulations, we opted for water as we tested out the end-to-end system on an open bench. 
An essential equipment in any laboratory is an accurate balance; here, we opted for the Fisher Scientific PPS4102 top pan balance.
Another key instrument is a hot plate and stirrer, where for this setup we used the IKA RCT Digital hot plate and stirrer. 

ARChemist was implemented using Python 3.8 and MongoDB Enterprise was used. 
All experiments were run using ROS Noetic on Ubuntu 20.04.

\begin{figure}
    \centering
    \subfigure[]{\includegraphics[width=0.29\textwidth]{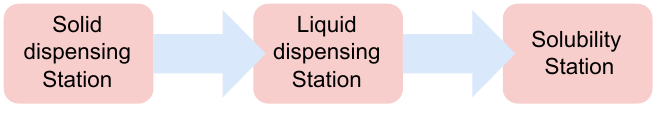} \label{fig:solubility_steps}} 
    \subfigure[]{\includegraphics[width=0.4\textwidth]{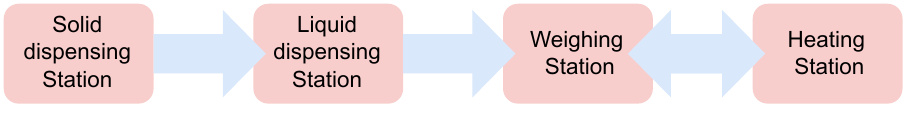} \label{fig:crystalization_steps}} 
    \caption{(a) Automated solubility screening workflow steps. (b) Automated crystallisation workflow steps.}
    \label{fig:workflows_steps}
\end{figure}

\subsection{Case Study: Automated Solubility Screening}
\label{subsec:case_study_1}

Solubility screening is the process of determining when the solute has dissolved in the solvent. 
This task is vital in material science research for solvent selection, in addition to being widely adopted in industries such as pharmaceuticals. 
To study this process using ARChemist, the Kuka KMR navigates to the solid dispensing system and puts a clean 10mL crimp top glass vial, with a magnetic stir bar, on the Quantos' carousel. 
% The Quantos is equipped with a dosing head containing the solid to be dispensed, which in our case is sodium chloride (salt). 
15mg of salt was dispensed and after this process finishes the Kuka KMR takes the vial and transports it to the Panda station. 
The Panda moves the vial to the liquid dispenser where the needle attached to the pump dispenses an initial dose of 2 mL solvent (water). 
Following this, the Panda moves the vial filled with solid and solvent to the stirrer plate, stirring the mixture for 1 second at 200 rpm.
Using an Intel RealSense D435i, RGB images are captured which will be used to compute the turbidity values~\cite{Zepel2020}, that would determine when the solute has dissolved.
The vial is then returned to the KUKA KMR for storage or disposal. 
This experiment was carried out 10 times with an average run time of 12 minutes. Compared to an experiment run by a human chemist, this is slower. However, this would free the chemist's time from doing repetitive and tedious tasks and focus on research. Moreover, this would allow to run experiments over long periods of time with no stoppage thus increasing the throughput. 
It ran successfully for 9 times and failed once due to Quantos QS3 timing out when taring, due to external vibrations. This experiment's steps are illustrated in Fig.~\ref{fig:solubility_steps}. 

\begin{center}
\begin{tabular}{ |c|c|c| } 
 \hline
 Number of runs & Success rate & Duration (minutes) \\ 
 \hline
 10 & 90\% & 12 \\ 
 \hline
\end{tabular}
\label{table1:case_study_1}
\end{center}
% https://pubs.acs.org/doi/pdf/10.1021/acs.accounts.0c00736

\subsection{Case Study: Crystallisation}
\label{subsec:case_study_2}

% The second study that we validated our architecture with was a crystallisation workflow.
Crystalline solids are useful in many applications, such as for semiconductors, in pharmaceutical formulations, and for molecular separations using porous crystals~\cite{Cui2019}. 
In our evaluation, we periodically heated and weighed the solution as it was assumed that crystallisation was complete when the mass stopped changing. 
Our workflow starts from the KUKA KMR repeats the same process of moving the clean vial, this time without a magnetic stir bar, to the Quantos. 
Here, 200 mg of sodium chloride is dispensed, followed by moving the vial to the Panda station, where the Panda moves the vial again to the liquid dispenser to dispense the solvent (2 mL of water).
The robot then records the mass of the vial with the solute and solvent by placing the sample on the scale.
For a fixed duration of time, the robot heats the solution at 60 degrees Celsius and continuously monitors the change in mass.
The crystallisation is deemed complete when the mass stabilises, that is, all of the solvent is lost. This experiment was carried out 5 times with an average run time of 2 hours and 10 minutes. Similar to the previous experiment, this saves time for the human chemist from repetitive and tedious tasks and increase throughput. 
The failed run was due to the Panda misplacing the vial when moving it to the scale.
Fig~\ref{fig:crystalization_steps} illustrates this experiment's steps.\par

Fig~\ref{fig:workflows_screenshot} shows a screenshot of the robots in operation in the crystallisation workflow. The accompanying video (link: \href{https://youtu.be/mcEYyOBjKpU}{https://youtu.be/mcEYyOBjKpU}) showcases the system performing both experiments. 

\begin{center}
\begin{tabular}{ |c|c|c| } 
 \hline
 Number of runs & Success rate & Duration (minutes) \\ 
 \hline
 5 & 80\% & 130 \\ 
 \hline
\end{tabular}
\label{table2:case_study_2}
\end{center}

%% file: section/conclusion.tex
\section{CONCLUSION}
\label{sec:conclusion}

This paper presents ARChemist, the first reconfigurable architecture, based on open-source tools such as ROS, for chemistry lab automation that uses widely available commercial robot platforms and digital laboratory instruments. 
In close collaboration with chemists, we developed a system that allows scientists to program their autonomous experimental setup using intuitive chemical recipes. 
To demonstrate its application to real-world experiments, we presented the architecture in two case studies: solubility screening and crystallisation.
These experiments illustrated how heterogeneous robots could work together in a material chemistry laboratory to relieve the human scientist from repetitive tasks. 
As this architecture has demonstrated its excellent potential for helping laboratory-based scientists to accelerate their daily experiments, we plan on extending it with recent advances in machine learning~\cite{Christensen2021}, which have shown much success in material discovery, which is our application domain. 
This would bring us closer to our long term goal of having multiple robotic scientists collaboratively working to generate and test hypotheses with minimal researcher intervention
We believe that this work makes us progress towards this vision.

\begin{figure}
    \centering
    \includegraphics[width=0.49\textwidth]{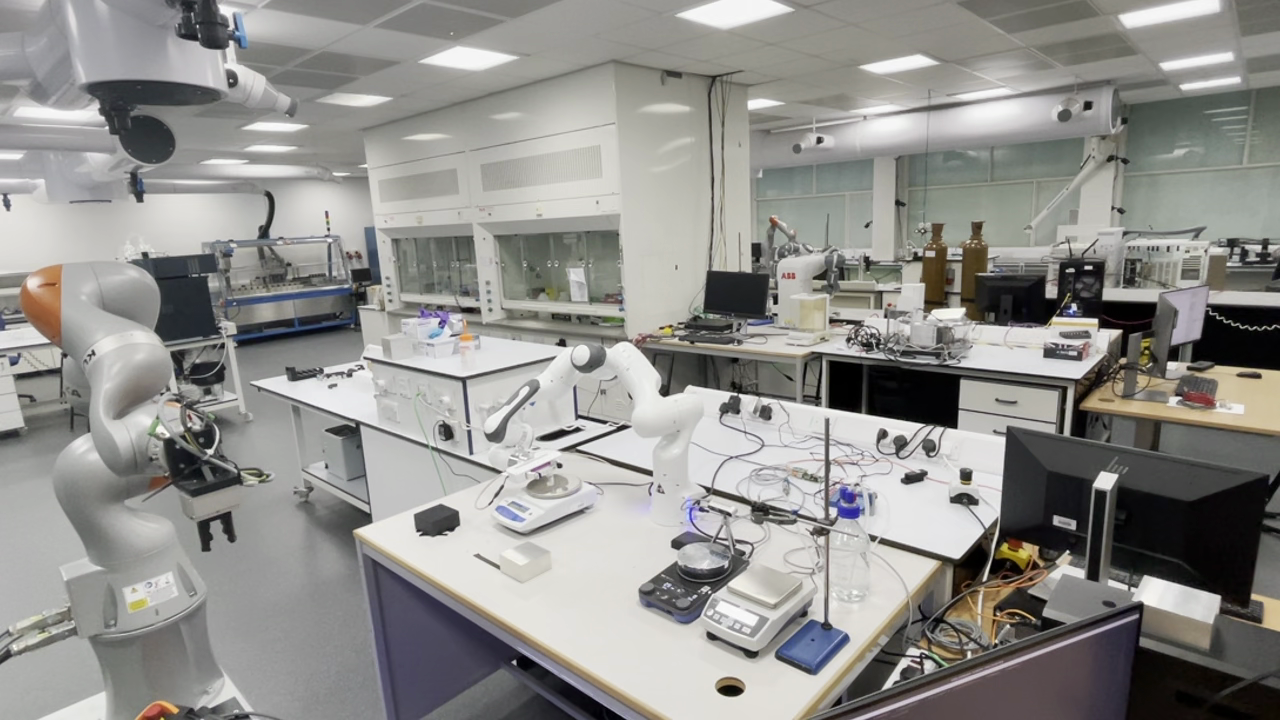}
    \caption{The Kuka KMR and Panda arm in operation in the automated crystallisation workflow.}
    \label{fig:workflows_screenshot}
\end{figure}